\documentclass[10pt,twocolumn,letterpaper]{article}

\PassOptionsToPackage{table}{xcolor}

\usepackage[pagenumbers]{cvpr} 

\usepackage{microtype}

\usepackage[utf8]{inputenc} 
\usepackage[T1]{fontenc}    
\usepackage{url}            
\usepackage{booktabs}       
\usepackage{amsfonts}       
\usepackage{nicefrac}       
\usepackage{amsmath}
\usepackage{graphicx}
\usepackage{pifont}         

\definecolor{cvprblue}{rgb}{0.21,0.49,0.74}
\usepackage[pagebackref,breaklinks,colorlinks,allcolors=cvprblue]{hyperref}

\def\confName{CVPR}
\def\confYear{2027}

\title{CRF Loss is How Networks Should Learn Boundaries \\ in Weakly Supervised Segmentation}

\author{
Joshua Li ~~~~~~~~~~ Yuri Boykov \\
Cheriton School of Computer Science \\
University of Waterloo \\
{\tt\small \{j234li,yboykov\}@uwaterloo.ca}
}

\begin{document}
\maketitle
\begin{abstract}
  Weakly Supervised Semantic Segmentation (WSSS) learns pixel-level predictions from image-level tags. Recent work focuses on improving coarse CAMs extracted from large vision-language models (commonly CLIP), but does little to improve their accuracy along segment boundaries. That job is instead delegated to a post-processing method like DenseCRF. However, because DenseCRF relies on low-level colour cues, it can flip correct labels to incorrect ones when neighbouring pixels share similar colours. SAM has recently been adopted as a natural alternative, yet it simply takes on DenseCRF's role as an intermediate "refinement" step that outputs one-hot pseudo-labels in prior work. By discarding the valuable uncertainty in CAMs, these one-hot pseudo-labels turn borderline errors into confidently wrong targets. Our key insight is that CAMs should supervise training alongside SAM boundaries, each through its own loss, rather than being fused together into a single hard target. Inspired by CRF potentials, we propose a framework that disentangles soft pseudo-labels as unary supervision and binary edge maps as pairwise supervision. We realize our framework in a single-stage model, DS-CRF, using CAMs from \texttt{dino.txt} and boundaries from SAM. DS-CRF sets a new state-of-the-art of 56.5\% mIoU on MS COCO among single-stage methods.
\end{abstract}

\section{Introduction}
\label{sec:intro}

\begin{figure*}[t]
  \centering
   \includegraphics[width=0.75\linewidth, trim=34.65bp 14.85bp 15.75bp 22.5bp, clip]{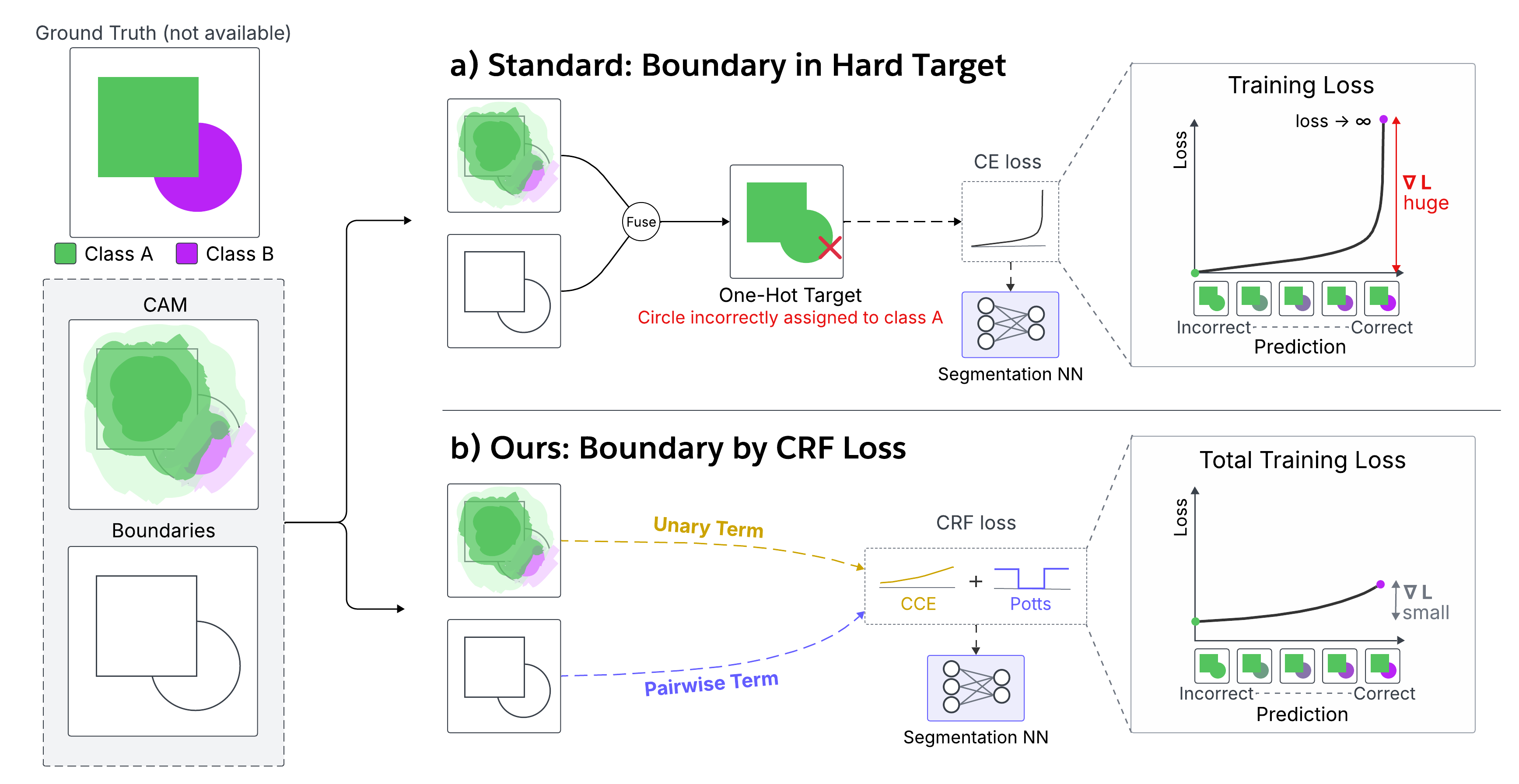}

   \caption{\textbf{Boundary in hard target vs. boundary by CRF loss.} The CAM for class A leaks into the circle, which belongs to class B. \textbf{(a)} Prior work uses DenseCRF or SAM to fuse the CAM and the boundaries into a one-hot (hard) target that assigns the whole circle to class A, so the correct prediction incurs infinite loss. \textbf{(b)} Our CRF loss disentangles the two signals, using soft, CAM-derived pseudo-labels as unary supervision and boundaries as pairwise supervision. Since no class is ever hard-imposed, the correct prediction is penalized far less.}
   \label{fig:motivation}
\end{figure*}

Semantic segmentation predicts a class label for each pixel in an image. Unlike fully supervised methods that rely on dense pixel-level annotations, Weakly Supervised Semantic Segmentation (WSSS) leverages cheaper forms of supervision, such as scribbles \cite{zhang_soft_self_labeling}, bounding boxes \cite{kulharia_box2seg}, or image-level tags \cite{ahn_affinity_net, wei_adversarial_erasing, lin_clip_es}. Among these, image-level tags are especially popular due to their wide availability in web-scale datasets as captions, and will be our focus in this work.

Unsurprisingly, the central question in WSSS has been how to turn image-level tags into dense pixel-level labels. Once such labels exist, training a segmentation network on them with cross-entropy is straightforward, since the problem is reduced to fully supervised segmentation. The labels usually start as Class Activation Maps (CAMs) from a CNN or, increasingly, from a CLIP ViT. However, CNN CAMs are known to exhibit imprecise boundaries, and ViT CAMs are limited by the patch resolution (e.g., $16\times16$ pixels). As such, a common practice is to refine CAM boundaries with DenseCRF \cite{krahenbuhl_densecrf} post-processing and argmax them into one-hot pseudo-labels. Some recent works replace DenseCRF with SAM \cite{kirillov_sam}, for example by labelling each SAM mask with the class whose CAM it overlaps most. While SAM provides higher-level object boundaries, these methods still apply it as an intermediate refinement step and output one-hot pseudo-labels.

If CAMs were sufficiently accurate, this approach would have no issues. In practice \textbf{CAMs are never sufficiently accurate}, and "refining" them into one-hot pseudo-labels can amplify their errors in both size and confidence. DenseCRF adds errors of its own, since it relies on low-level appearance cues that need not align with semantic object boundaries. Segmentation networks trained on such confidently incorrect pseudo-labels are thus prone to reproduce the same errors themselves.

\Cref{fig:motivation} illustrates the problem. The CAM
\footnote{We define a CAM to be an unnormalized activation map, and pseudo-labels to be per-pixel vectors that form valid probability distributions (a special case of CAMs). We use the two terms interchangeably when the distinction does not matter.}
for class A spills into the circle, whose pixels belong to class B. Methods that use DenseCRF or SAM as a refinement step fuse these incorrect CAMs with the boundaries and argmax the result into a one-hot target, assigning the entire circle to class A. A segmentation network trained on this target is encouraged to reproduce the error, since predicting the correct ground truth segmentation would incur infinite cross-entropy loss. Note that in the figure we draw SAM-like boundaries for DenseCRF as well, where they can instead be read as colour-coherent regions.

We argue that these two forms of supervision should be \textbf{disentangled} into two terms of a Conditional Random Field (CRF) loss. Rather than folding boundaries into one-hot pseudo-labels, we supervise the segmentation network explicitly on a boundary term that encourages uniform predictions within a region, without prescribing which label that region should take. We further exploit the uncertainty information in CAMs by converting them into \textit{soft} rather than \textit{hard} (i.e. one-hot) pseudo-labels, which modulates the strength of the unary cross-entropy term so that pixels with uncertain pseudo-labels are penalized less than certain ones. In \cref{fig:motivation}, our network can assign the circle to class B at much lower loss, while still being pushed to give every pixel in it the same label.

Our framework builds on a quadratic relaxation of the CRF model. It extends the idea of \cite{tang_reg_losses}, which uses the CRF energy as a training objective rather than minimizing it over discrete labels at inference time (e.g. graph cut \cite{boykov_graph_cut} or DenseCRF \cite{krahenbuhl_densecrf}), making it compatible with standard deep learning. We make two further contributions relative to prior work in this area.

First, classical CRF formulations \cite{krahenbuhl_densecrf, grady_random_walks, boykov_graph_cut} derive continuous pairwise affinities $w_{i,j}$ from low-level colour differences. To exploit high-level SAM masks, we instead derive binary pairwise affinities from discrete boundaries. However, directly converting boundaries into binary affinities leads to poor model convergence in practice. We hypothesize that this is because image quantization creates ambiguous pixels near object boundaries that cannot be definitively assigned to a class. We refer to this effect as \textit{partial voluming}, borrowing terminology from 3D medical imaging. We address this with a simple \textit{dilation} (or widening) of the boundary edge maps. While our binary affinities also accommodate superpixel boundaries, we achieve the best results with high-level SAM boundaries.

Second, we replace standard cross-entropy with the more robust collision cross-entropy \cite{zhang_soft_self_labeling}. While \cite{tang_reg_losses} assumes access to accurate ground truth scribbles, tag-derived CAMs are much more inaccurate. Collision cross-entropy accounts for this by weighting the loss by pseudo-label confidence, allowing the model to learn from soft pseudo-labels without treating them as definitive targets. In contrast to \cite{zhang_soft_self_labeling}, which jointly optimizes the pseudo-labels alongside the network, we apply collision cross-entropy to fixed pseudo-labels and thus simplify the training process.

We realize our framework in a single-stage method, DS-CRF, built around a DINOv3 backbone with a \texttt{dino.txt} head for CAM generation and a separate SAM branch for boundary generation. DS-CRF achieves state-of-the-art WSSS performance among single-stage methods, including a new best of 56.5\% mIoU on MS COCO \cite{lin_ms_coco}.

\section{Related Works and Preliminaries}
\label{sec:related}

\subsection{Weakly Supervised Semantic Segmentation}

WSSS approaches typically fall into multi- and single-stage paradigms. Multi-stage approaches follow a sequential pipeline: (1) generating Class Activation Maps (CAMs) \cite{zhou_cam} from a classification network, (2) refining these CAMs into pseudo-labels, and (3) training a fully supervised segmentation model using the pseudo-labels as ground truth. In contrast, traditional single-stage approaches are trained with a unified model containing both a classification head and a segmentation head. While more streamlined, single-stage approaches typically underperform multi-stage approaches.

Regardless of the paradigm, most WSSS methods train a segmentation model with cross-entropy using refined pseudo-labels as targets. Let $\Omega$ denote the set of image pixels and $C$ the number of classes (including background). Given a discrete pseudo-label $\bar{y}_i \in \{1, \dots, C\}$ at pixel $i \in \Omega$, the per-pixel loss is the \textit{negative log-likelihood} (NLL)
\begin{equation}
    -\ln \sigma_i^{\bar{y}_i},
\label{eq:nll}
\end{equation}
where $\sigma_i \in \Delta^{C}$ is the trained model prediction (i.e. a categorical distribution from softmax). Here,
$\Delta^{C} = \{(p^1,\dots,p^C) \mid p^c \ge 0, \sum_{c=1}^{C} p^c = 1\}$
denotes the $C$-class probability simplex. If $\bar{y}_i$ is represented as a \textit{one-hot} distribution $y_i = (y_i^1,\dots,y_i^C)\in \Delta^{C}_{\{0, 1\}}$  such that $y_i^c=[c=\bar{y}_i]\in\{0,1\}$ for the Iverson bracket $[ \cdot ]$, then NLL \eqref{eq:nll} is equivalent to the \textit{standard cross-entropy}
\begin{equation}
    H_{\text{CE}}(y_i, \sigma_i) = -\sum_{c} y_i^c \ln \sigma_i^c.
\label{eq:cross_entropy}
\end{equation}
This formulation naturally extends to soft pseudo-labels $y_i\in\Delta^{C}$, which we discuss together with cross-entropy variants in \cref{sec:method_unary}.

Early research in WSSS explored pixel-wise affinities \cite{ahn_affinity_net, fan_cian}, adversarial erasing \cite{wei_adversarial_erasing, hou_seenet}, and saliency maps as auxiliary supervision \cite{wei_dilated_wsss, wei_adversarial_erasing}. More recently, the field has progressed by integrating large pre-trained models like CLIP \cite{radford_clip}, DINO \cite{oquab_dino_v2}, and SAM \cite{kirillov_sam}. Given that WSSS relies on image-text modalities, CLIP has become the most prevalent choice. For instance, CLIMS \cite{xie_clims} leverages CLIP to ensure target region completeness and background suppression in CAMs, while CLIP-ES \cite{lin_clip_es} introduces a Softmax-GradCAM that produces CAMs directly from CLIP's image encoder. WeCLIP \cite{zhang_weclip} builds on CLIP-ES by adopting a single-stage framework with online affinity-based refinement.

In contrast, DINO has seen relatively limited use within WSSS. Existing methods such as ECA \cite{wu_eca} and WeCLIP+ \cite{zhang_weclip_plus} employ DINO as an auxiliary feature encoder rather than as a source of CAMs. We instead adopt \texttt{dino.txt} \cite{jose_dino_txt}, which extends DINO with additional training on image-caption pairs (similar to CLIP), and use it to generate CAMs directly. Its CAMs remain fairly inaccurate, but cover object extents more completely than Softmax-GradCAM \cite{lin_clip_es} on CLIP, likely thanks to the stronger features from DINO's image-only pre-training. As a result, fewer pixels in the object interior are misclassified with high confidence, which aligns better with our confidence-weighted unary loss.

SAM \cite{kirillov_sam} has recently become another powerful tool for WSSS, owing to its ability to produce accurate class-agnostic masks. Some approaches \cite{chen_sam_wsss, liu_co2sam} prompt SAM with bounding boxes obtained from open-vocabulary detectors \cite{liu_grounding_dino}, while others \cite{chen_sam_enhances} apply it as post-processing to one-hot pseudo-labels taken from prior WSSS methods. S2C \cite{kweon_s2c} offers a more principled framework, prompting SAM with the local maxima of CAMs, yet still hardens the result into one-hot pseudo-labels. These are produced online, supervising a CAM-generating network as it trains. Similarly, FMA-WSSS \cite{yang_fma_wsss} assigns each SAM mask the label of the CAM it overlaps most to produce one-hot pseudo-labels.

\begin{figure*}[t]
  \centering
   \includegraphics[width=0.7\linewidth, trim=14.85bp 20.25bp 14.85bp 22.95bp, clip]{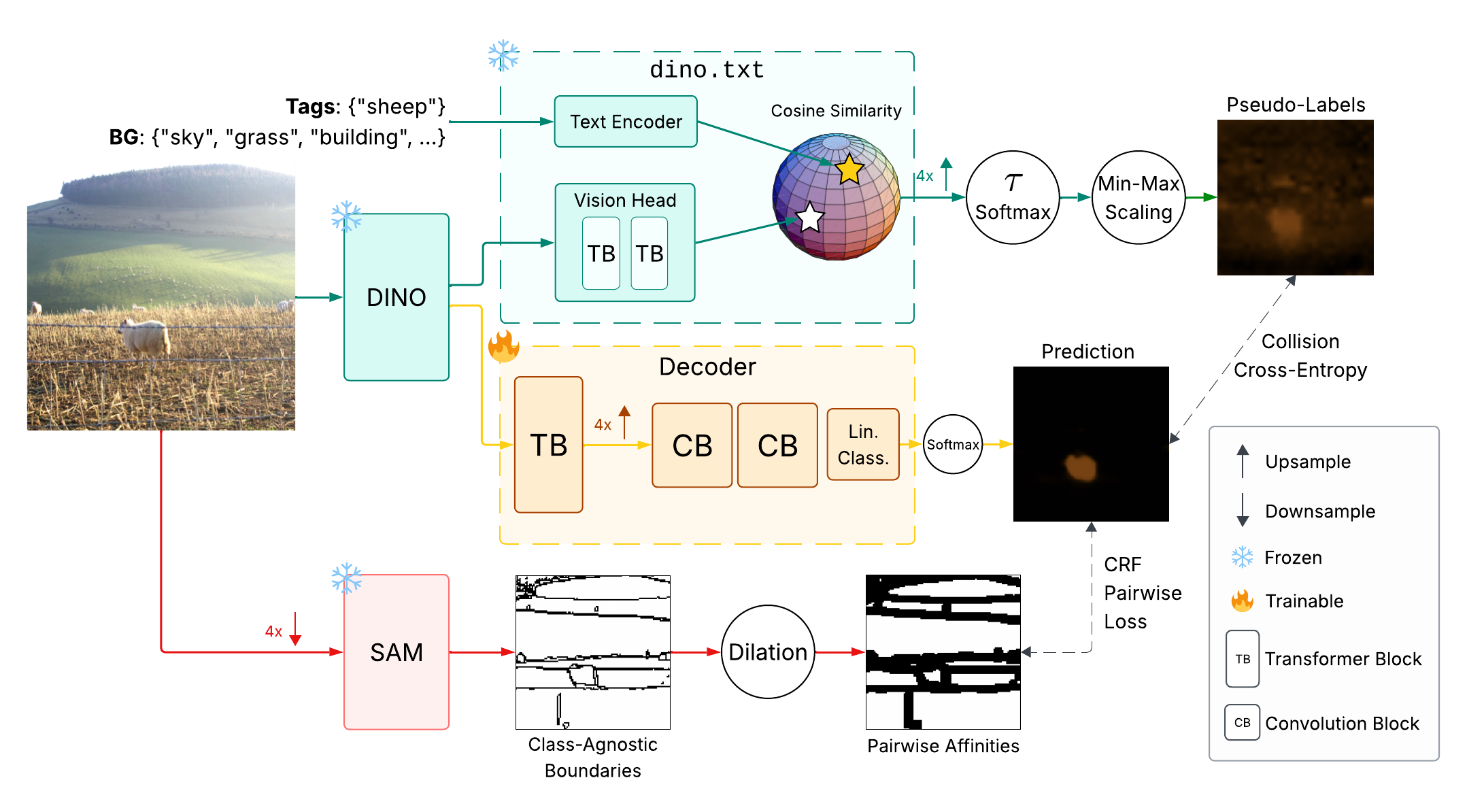}
   \caption{\textbf{DS-CRF architecture}. We design a lightweight decoder on top of the frozen DINO backbone, running in parallel with the \texttt{dino.txt} vision head. CAMs are generated by computing the cosine similarity between \texttt{dino.txt} patch and text embeddings, which serve as soft unary targets for collision cross-entropy after normalization. Separately, the image is passed through SAM to produce class-agnostic boundaries, then dilated to define pairwise affinities within our CRF pairwise loss. (Sphere image from \cite{weisstein_sphere}).}
   \label{fig:architecture}
\end{figure*}

\subsection{CRF Model}

Pairwise or higher-order CRF models have long been a staple in image segmentation and other computer vision literature \cite{Li:MRFBook,BlakeZis:87,boykov_graph_cut,krahenbuhl_densecrf, kohli_robust_pn, ladicky_hierarchical_crf, kolesnikov2016seed}. Among them, DenseCRF \cite{krahenbuhl_densecrf} is particularly influential in the context of semantic segmentation and WSSS. It is often used as a post-processing step, either to refine CAMs, as discussed earlier \cite{ahn_affinity_net, wang_seam, lin_clip_es, tang_cpal, zhao_psdpm}, or to clean up the final predictions of a segmentation network \cite{chen_deeplab,zhang_weclip, jang_dial}. With slight abuse of terminology (for an easier comparison with our approach later), we refer to both inputs as model predictions $\sigma_i$. To refine them, DenseCRF uses a Gaussian kernel over colour differences, guiding segmentation boundaries to high-resolution (but non-discriminative) intensity edges in the input image. Specifically, it uses mean-field inference to optimize the following energy over discrete variables $\bar{y}_i \in \{{1,\dots,C}\}$ representing refined class labels:
\begin{equation}
-\sum_{i \in \Omega} \ln \,\sigma_i^{\bar{y}_i} + \lambda \sum_{(i,j) \in \mathcal{N}} w_{i,j} \cdot [\bar{y}_i \ne \bar{y}_j],
\label{eq:crf_energy}
\end{equation}
where $\lambda$ is a weighting hyperparameter for the pairwise regularization term.  In \cref{eq:crf_energy}, the first term denotes the per-pixel or {\em unary} cost of assigning variable $\bar{y}_i$ to class $c$, given coarse prediction $\sigma_i=(\sigma_i^1,\dots,\sigma_i^C)$. The second term, known as the {\em Potts model} in CRF literature \cite{boykov_fast_approximate}, penalizes label discontinuities between pairs of pixels $(i,j) \in\mathcal{N}$ in a given neighbourhood $\mathcal{N}$, weighed by pairwise affinities $w_{i,j}$.

We have already seen that post-processing CAMs can amplify incorrect training signals. Post-processing final predictions also introduces errors that the network cannot recover from, as the pipeline is no longer end-to-end trainable. As such, there have been efforts to integrate CRF into model training itself. Some approaches incorporate CRF inference into the network architecture, e.g. using RNNs \cite{zheng_crf_rnn} or message estimator CNNs \cite{lin_message_passing}.
Other methods directly use CRF as a loss function for scribble-supervised training of segmentation networks \cite{tang_reg_losses}. 

In fact, using CRF as network loss represents a conceptual shift from using CRF as post-processing or as network architecture. It inspired our approach to WSSS with image-level tags. Rather than treating the given predictions $\sigma_i$ as coarse input and postprocessing it into refined discrete class labels $\bar{y}_i$, we directly CRF-regularize WSSS network output $\sigma_i$ during its training. To do so, the pairwise CRF loss is relaxed to operate on continuous predictions $\sigma_i \in \Delta^C$. We derive coarse pseudo-labels $y_i \in \Delta^C$ from CAMs without post-processing and use them as approximate targets. To train refined predictions $\sigma_i$, we adopt the quadratic relaxation \cite{grady_random_walks, zhang_soft_self_labeling} of the Potts regularization model:
\begin{equation}
\sum_{i \in \Omega} H(y_i, \sigma_i) + \lambda \sum_{(i,j) \in \mathcal{N}} w_{i,j} \cdot\frac{1}{2} \lVert \sigma_i-\sigma_j \rVert^2,
\label{eq:quad_crf}
\end{equation}
where $H$ denotes a cross-entropy loss (e.g. $H_\text{CE}$ in \cref{eq:cross_entropy}) and $\lVert \cdot \rVert^2$ is the squared $L_2$ norm. This objective can be optimized via standard gradient descent.

Within the Potts model, the neighbourhood $\mathcal{N}$ can be defined in several ways, including as the nearest-neighbour grid \cite{grady_random_walks, boykov_graph_cut} or fully connected graph \cite{krahenbuhl_densecrf}. Fully connected graphs capture longer-range interactions but require specialized implementations like fast Gaussian filtering \cite{krahenbuhl_densecrf, tang_reg_losses}. In contrast, nearest-neighbour grids limit interactions to local pixels, resulting in a much simpler implementation. We use the 4-connected nearest-neighbour grid and show that it provides sufficient regularization.

Given a neighbourhood, the pairwise affinity $w_{i,j}$ is typically designed to encourage label consistency between pixels with similar low-level appearances. Let $I_i$ and $I_j$ denote the colour vectors at pixels $i$ and $j$. A common choice \cite{grady_random_walks, boykov_graph_cut, krahenbuhl_densecrf} is the Gaussian kernel over their colour difference:
\begin{equation}
    w_{i,j}=\exp \left( -\frac{\lVert I_i -I_j \rVert^2}{2 \delta^2  } \right).
\label{eq:pairwise_gaussian}
\end{equation}
This is used in both nearest-neighbour and fully connected settings, sometimes augmented with additional term(s) based on positional difference. It weakens regularization across high-contrast intensity edges, aligning predictions with low-level contours and improving boundary accuracy.

\section{Method}

\subsection{Architecture}

Our architecture consists of four main components: the DINO backbone, the \texttt{dino.txt} vision head and text encoder, the SAM model, and a lightweight segmentation decoder (see \cref{fig:architecture}). The first three components are kept frozen, and only the decoder is trained. We reuse rich features from the DINO backbone for two purposes: CAM generation through \texttt{dino.txt}, and segmentation prediction through our decoder. In this section, we briefly describe the DINO backbone and the decoder.

Given an input image $\mathbf{I} \in \mathbb{R}^{3 \times H \times W}$, the DINO backbone outputs patch tokens $\mathbf{P} \in \mathbb{R}^{D \times \frac{HW}{256}}$ (using a patch size of $16 \times 16$), four register tokens and a \texttt{CLS} token. These are passed to the decoder, which consists of a Transformer Block (TB), two Convolution Blocks (CB) and a linear classifier.

The transformer block first jointly processes all tokens. We then remove the register and \texttt{CLS} tokens, reshape the patch tokens, and bilinearly upsample them by a factor of $4$ to obtain a feature map of shape $D \times \frac{H}{4} \times \frac{W}{4}$. This spatial resolution enables more precise boundary regularization while being computationally cheaper than full resolution. The upsampled features are subsequently refined by the two convolution blocks, each implemented as a ResNet basic block with two $3 \times 3$ convolutions and a residual connection. A linear classifier projects the features to $C$ channels (including background), and we apply a final softmax to yield segmentation predictions $\boldsymbol{\sigma} \in (\Delta^C)^{\frac{H}{4} \times \frac{W}{4}}$.

\subsection{Pseudo-labels and Cross-Entropy Variants}
\label{sec:method_unary}

To generate CAMs, the DINO patch, register, and \texttt{CLS} tokens are also passed through two transformer blocks within the frozen \texttt{dino.txt} vision head. Similar to the decoder, we only use the output patch tokens $\mathbf{Z} \in \mathbb{R}^{D \times \frac{HW}{256}}$ and discard the rest.

We turn to the \texttt{dino.txt} text encoder to generate text embeddings from natural language prompts. Following CLIP-ES \cite{lin_clip_es}, we define a set of classes $\{\mathcal{T} \cup \mathcal{B}\}$, where $\mathcal{T}$ contains image-specific ground truth tags (such as \textit{boat} or \textit{cat}) and $\mathcal{B}$ contains predefined background classes (such as \textit{grass} and \textit{sky}). We also use common techniques like class name optimization \cite{jose_dino_txt} and prompt ensembling. The text encoder processes prompts to return text embeddings $\mathbf{T} \in \mathbb{R}^{D \times (|\mathcal{T}| + |\mathcal{B}|)}$.  

We normalize both the patch and text embeddings along the feature dimension $D$, then compute their cosine similarities as
\begin{equation}
\mathbf{S}' = \mathbf{T}^\top \mathbf{Z},
\end{equation}
which is reshaped to obtain $\mathbf{S}' \in \mathbb{R}^{(|\mathcal{T}| + |\mathcal{B}|) \times \frac{H}{16} \times \frac{W}{16}}$. We take the maximum similarity across background set $\mathcal{B}$ at every pixel to produce a single background channel. After bilinear upsampling, this arrives at our CAM $\mathbf{S} \in \mathbb{R}^{(|\mathcal{T}|+1)  \times\frac{H}{4} \times \frac{W}{4}}$, which can then be processed into pseudo-labels.

\begin{figure}[t]
   \centering
   \includegraphics[width=0.8\linewidth, trim=15.3bp 22.5bp 15.3bp 15.3bp, clip]{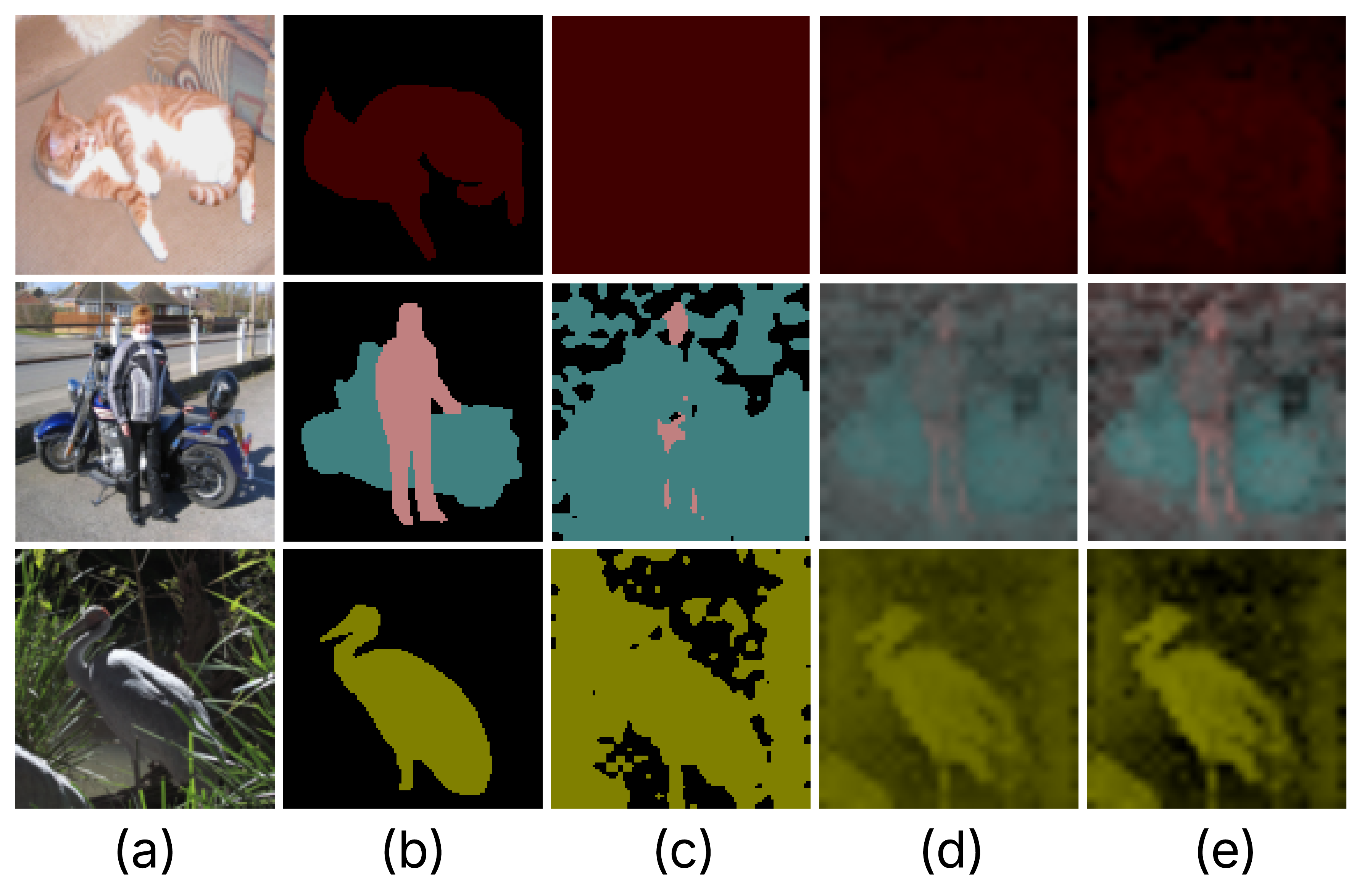}
   \caption{Pseudo-labels from \texttt{dino.txt} CAMs. (a) Image. (b) Ground truth. (c) Hard pseudo-labels. (d) Soft pseudo-labels. (e) Soft pseudo-labels with min-max scaling.}
   \label{fig:pseudo_labels}
\end{figure}

\paragraph{Standard Cross-Entropy.}
The de facto way to supervise segmentation training is cross-entropy on one-hot pseudo-labels. For each pixel $i$, we take the argmax of the CAM $\mathbf{S}$ over classes to obtain a hard target $\bar{y}_i \in \{1,\dots,C\}$, which can be substituted directly into the NLL loss in \cref{eq:nll} to train the decoder's pixel-level predictions $\sigma_i$.

However, as \cref{fig:pseudo_labels}(c) shows, these hard pseudo-labels contain many false positives, i.e. background pixels assigned to a foreground class. Because NLL loss treats every label as equally certain, the decoder is pushed to fit these errors as strongly as the correct labels, which leads to highly inaccurate segmentation.

We argue that CAM uncertainty is a valuable signal for supervision strength that should be retained rather than discarded. As such, we move away from hard (i.e. one-hot) pseudo-labels in favour of soft distributions. We transform CAMs into probability distributions using temperature-scaled softmax
\begin{equation}
y_i^c = \frac{\exp(\mathbf{S}_i^c / \tau)}{\sum_{k \in \mathcal{T}^{+1}} \exp(\mathbf{S}_i^k / \tau)} \quad \quad \quad \text{for}\ c \in \mathcal{T}^{+1},
\label{eq:soft_prob}
\end{equation}
where $\tau$ controls the sharpness of the distribution and $c$ ranges only over the tag classes $\mathcal{T}$ plus background, i.e. $\mathcal{T}^{+1}$. We notice that small $\tau$ values approximate argmax and thus amplify error as before, while large $\tau$ values overly smooth the distribution and prevent effective learning. Ideally, a pseudo-label should be sharp when it is correct and close to uniform when it is incorrect.

To achieve this, we fix a relatively large $\tau$ to maintain softness and apply a simple min-max scaling across the spatial dimension (similar to \cite{lin_clip_es, kweon_s2c}). Our motivation is that for every present tag (plus background), the corresponding pseudo-label map should correctly identify at least one pixel that belongs to the class and one that does not. Concretely, we apply the following operation after softmax:
\begin{equation}
    y_i^c = \frac{y_i^c - \min_{j}(y^c_j)}{\max_{j}(y^c_j) - \min_{j}(y^c_j)} \quad \quad \quad\text{for}\ c \in \mathcal{T}^{+1}.
\end{equation}
Following this scaling, we re-normalize the vector at each pixel $i$ and pad non-present classes with probability $0$ to obtain pseudo-labels $y_i \in \Delta^{C}$. Effectively, min-max scaling allows us to sharpen the distribution of pseudo-labels likely to be correct while preserving uncertainty in others. This is illustrated qualitatively in \cref{fig:pseudo_labels}(d) and (e). 

We can now use $y_i$ as a soft target in the standard cross-entropy loss defined in \cref{eq:cross_entropy}. However, doing so simply propagates pseudo-label uncertainty to the segmentation output. This becomes clear from the decomposition
\begin{equation}
    H_{\text{CE}}(y_i, \sigma_i)=KL(y_i \Vert \sigma_i)+H(y_i),
\label{eq:ce_expanded}
\end{equation}
where $KL$ denotes KL-divergence and $H$ is the standard entropy. Since the entropy term is constant with respect to $\sigma_i$, minimizing the loss reduces to minimizing KL divergence, which is uniquely $0$ when $\sigma_i=y_i$. Evidently, $\sigma_i$ will be uncertain if the pseudo-label $y_i$ is uncertain. Moreover, because the quadratic relaxation used in the pairwise loss is not tight, the model is never explicitly encouraged to produce hard outputs. As a result, training with standard cross-entropy on soft pseudo-labels yields uncertain segmentation predictions.

\paragraph{Collision Cross-Entropy.}
Instead, we adopt collision cross-entropy \cite{zhang_soft_self_labeling} with soft pseudo-labels, defined as
\begin{equation}
    H_{\text{CCE}}(y_i, \sigma_i) = - \ln \sum_{c} y_i^c\sigma_i^c.
\end{equation}
This can equivalently be expressed as
\begin{equation}
    H_{\text{CCE}}(y_i, \sigma_i) = - \ln cos(y_i, \sigma_i)+ \frac{H_2(y_i)+H_2(\sigma_i)}{2},
\label{cce_expanded}
\end{equation}
where $cos(y_i, \sigma_i)$ denotes the cosine similarity between vectors $y_i$ and $\sigma_i$, and $H_2$ is the second-order Rényi entropy.

Collision cross-entropy offers two advantages. First, in contrast to \cref{eq:ce_expanded}, collision cross-entropy contains an entropy term $H_2(\sigma_i)$ which encourages hard predictions. Second, its gradient scales with the confidence of the pseudo-label: sharper pseudo-labels induce stronger gradients in the loss landscape, while more uniform pseudo-labels produce weaker ones. In the limiting cases, the loss reduces to NLL when $y_i$ is one-hot, and a constant when $y_i$ is uniform. We argue that this behaviour enables the model to prioritize learning from confident pseudo-labels, which are more likely to be correct, while remaining flexible towards uncertain pseudo-labels.

\subsection{Pairwise Regularization}
\label{sec:method_pairwise}
We now consider the pairwise term that encourages boundary regularization in \cref{eq:quad_crf}. We replace its single weighting parameter $\lambda$ with class-specific weights $\lambda^c$, which gives our final CRF loss:
\begin{equation} \mathcal{L}_{\text{CRF}}= \sum_{i \in \Omega} H(y_i, \sigma_i) + \sum_{(i,j) \in \mathcal{N}} w_{i,j} \cdot \sum_c \frac{\lambda^c}{2} (\sigma_i^c-\sigma_j^c)^2.
\label{eq:class_weighted_pairwise}
\end{equation}
This pairwise term enables different regularization strengths for different classes. Some classes (e.g., \textit{cat}, \textit{aeroplane}) benefit from a higher $\lambda^c$. However, for other classes, strong regularization can cause the pairwise term to dominate the unary term, leading to degenerate solutions where predictions collapse to a single class (typically the background). In such cases, lower values for $\lambda^c$ are desirable.

\begin{figure}[t]
   \centering
   \includegraphics[width=0.55\linewidth, trim=14.85bp 15.3bp 14.85bp 14.85bp, clip]{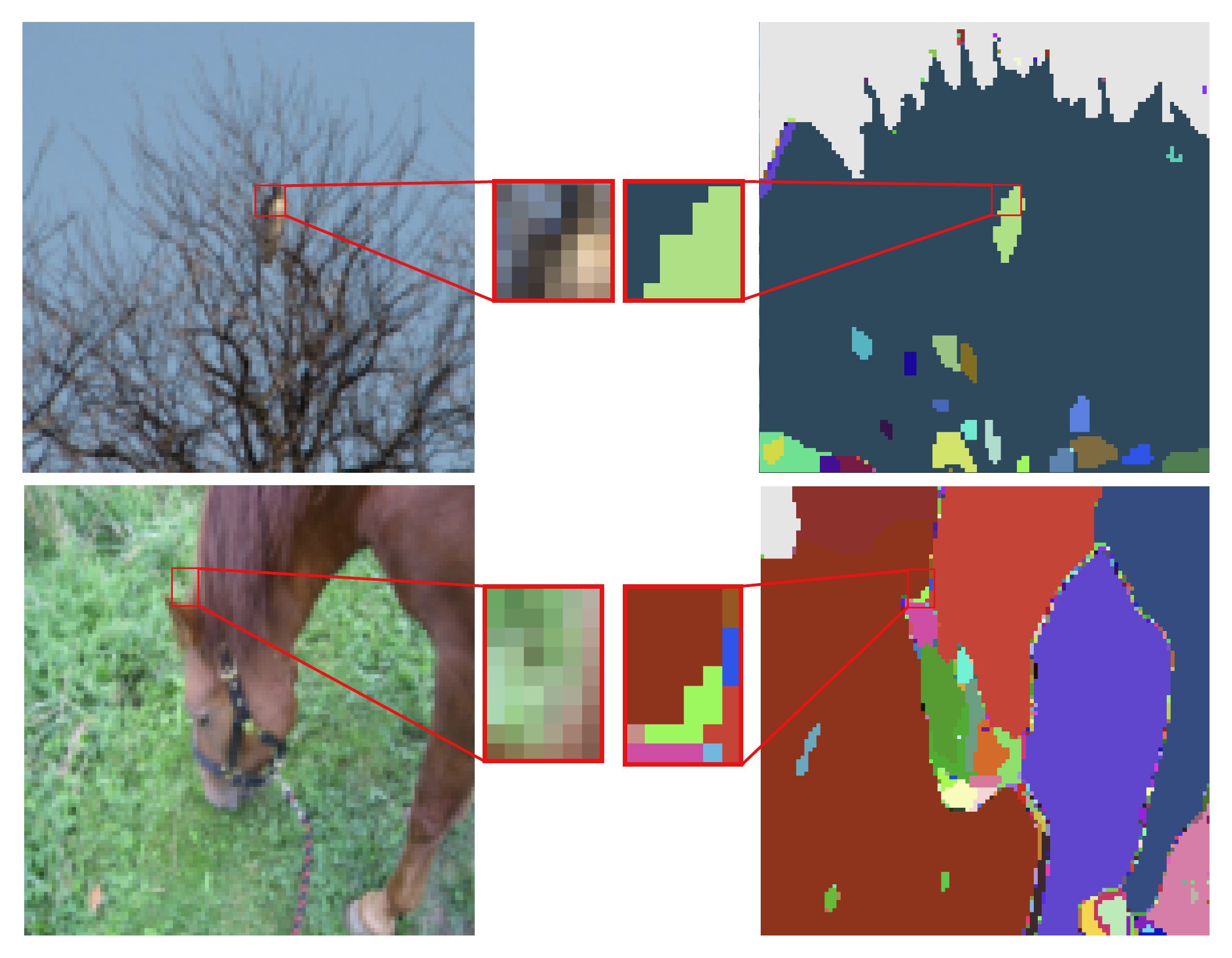}
   \caption{Effect of partial voluming on SAM superpixels. \textit{Left}: input images. \textit{Right}: SAM superpixels from aggregating SAM boundaries.}
   \label{fig:partial_voluming}
\end{figure}
To compute the pairwise affinities $w_{i, j}$ between adjacent pixels $i$ and $j$ on a 4-connected grid, we use high-level boundaries extracted from SAM. Specifically, we apply SAM’s automatic mask generator to a $4\times$ downsampled version of the input image to produce a set of $N$ binary masks
\begin{equation}
    \mathcal{M}=\{M^1,\dots,M^N\},
\end{equation}
where each mask is $M^n \in \{0,1\}^{\frac{H}{4} \times \frac{W}{4}}$. We then define the binary pairwise affinity as
\begin{equation}
w_{i,j}=\begin{cases}
      0, & \text{if}\ M_i^{n} \ne  M_j^{n}\ \text{for any}\ n\\
      1, & \text{otherwise}.
    \end{cases}
\label{eq:pairwise_affinity}
\end{equation}
That is, $w_{i,j}=0$ iff a boundary exists between pixels $i$ and $j$ on any produced mask. This ensures that the regularization term in \cref{eq:class_weighted_pairwise} only enforces prediction consistency within mask interiors. Along the boundaries, the term evaluates to $0$ and thus predictions are free to perform sharp class transitions. Importantly, to account for possible over-segmentation of the image, our loss acts as a permissive guide that allows for transitions at SAM boundaries without actively encouraging them. We also note that our binary $w_{i,j}$ is markedly different from the standard Gaussian kernel defined in \cref{eq:pairwise_gaussian}.

In practice, however, this naive implementation struggles. We attribute this to the partial voluming effect at object boundaries. Because boundary pixels often contain a mixture of the object and the background, there is no single "correct" way to draw a boundary through them. SAM tends to group these ambiguous pixels arbitrarily (see \cref{fig:partial_voluming}), capturing only specific combinations rather than allowing boundaries to be drawn around each individual pixel. As a result, the model receives inconsistent training signals across different images that prevent it from learning a generalized, sharp class transition at semantic boundaries.

To resolve this, we propose a simple dilation operation on the pairwise affinities to artificially widen the area where transitions are permitted. We achieve this by decomposing the affinities $w_{i,j}$ into vertical and horizontal binary edge maps $E_v \in \{0,1 \}^{(\frac{H}{4}-1) \times \frac{W}{4}}$ and $E_h \in \{0,1 \}^{\frac{H}{4} \times (\frac{W}{4}-1)}$. We define them as
\begin{equation}
   E_v(x,y) = w_{(x,y),(x+1,y)} , \quad E_h(x,y) =  w_{(x,y),(x,y+1)},
\end{equation}
where pixel indices in $w_{i,j}$ are replaced by the corresponding coordinates in the image, $i \Leftrightarrow (x,y)$. We then apply morphological dilation to both edge maps:
\begin{equation}
    E_v^d = E_v \oplus B^d, \quad E_h^d = E_h \oplus B^d
    \label{eq:dilation}
\end{equation}
where $B$ is a $d \times d$ square structuring element, dependent on dilation size $d \in \mathbb{N}$. Unlike conventional dilation, which expands regions of ones, we dilate over zeroes as these represent edges. Finally, we recombine $E_v^d$ and $E_h^d$ back into affinities $w^d_{i,j}$ for use in \cref{eq:class_weighted_pairwise}. By dilating these affinities, we effectively expand the boundaries, enabling the model to more reliably learn sharp class transitions.
\section{Experiments}

\begin{table*}[t]
\centering
\caption{Comparison with state-of-the-art WSSS methods. We report mIoU (\%) on the PASCAL VOC 2012 and MS COCO 2014 datasets. Supervision ($\text{Sup.}$) types are defined as: $\mathcal{I}$ (image-level labels), $\mathcal{L}$ (language), and $\mathcal{S}$ (SAM masks). For multi-stage methods, the backbone refers to the final segmentation model.}
\label{tab:sota}
\small
\begin{tabular}{l c c c c c c}
\toprule
Method & dCRF & Sup. & Backbone & VOC \textit{val} & VOC \textit{test} & COCO \textit{val}\\

\midrule
\multicolumn{7}{c}{Multi-Stage WSSS Methods} \\
KTSE$_{\text{ECCV'24}}$ \cite{chen_ktse} & \ding{51} & $\mathcal{I}$ & RN101 & 73.0 & 72.9 & 45.9 \\
MuP-VSS$_{\text{CVPR'25}}$ \cite{duan_mup_vss} & \ding{51} & $\mathcal{I}$ & WRN38 & 73.6 & 74.7 & 46.6 \\
CLIMS$_{\text{CVPR'22}}$ \cite{xie_clims} & \ding{51} & $\mathcal{I}+\mathcal{L}$ & RN101 & 70.4 & 70.0 & -- \\
CLIP-ES$_{\text{CVPR'23}}$ \cite{lin_clip_es} & \ding{51} & $\mathcal{I}+\mathcal{L}$ & RN101 & 73.8 & 73.9 & 45.4 \\
MMCST$_{\text{CVPR'23}}$ \cite{xu_mmcst} & \ding{51} & $\mathcal{I}+\mathcal{L}$ & WRN38 & 72.2 & 72.2 & 45.9 \\
CPAL$_{\text{CVPR'24}}$ \cite{tang_cpal} & \ding{51} & $\mathcal{I}+\mathcal{L}$ & RN101 & 74.5 & 74.7 & 46.8 \\
PSDPM$_{\text{CVPR'24}}$ \cite{zhao_psdpm} & \ding{51} & $\mathcal{I}+\mathcal{L}$ & RN101 & 74.1 & 74.9 & 47.2 \\
POT$_{\text{CVPR'25}}$ \cite{wang_pot} & \ding{51} & $\mathcal{I}+\mathcal{L}$ & RN101 & 76.1 & 76.7 & 47.9 \\

\midrule
\multicolumn{7}{c}{Single-Stage WSSS Methods} \\
DuPL$_{\text{CVPR'24}}$ \cite{wu_dupl} & \ding{51} & $\mathcal{I}$ & ViT-B & 73.3 & 72.8 & 44.6 \\
PCRE$_{\text{CVPR'25}}$ \cite{xu_pcre} & \ding{51} & $\mathcal{I}$ & ViT-B & 75.5 & 75.9 & 47.2 \\
DIAL$_{\text{ECCV'24}}$ \cite{jang_dial} & \ding{51} & $\mathcal{I}+\mathcal{L}$ & ViT-B & 74.5 & 74.9 & 44.4 \\
WeCLIP$_{\text{CVPR'24}}$ \cite{zhang_weclip} & \ding{51} & $\mathcal{I}+\mathcal{L}$ & CLIP & 76.4 & 77.2 & 47.1 \\
ExCEL$_{\text{CVPR'25}}$ \cite{yang_excel} & \ding{51} & $\mathcal{I}+\mathcal{L}$ & CLIP & 78.4 & 78.5 & 50.3 \\

\rowcolor{gray!15}
DS-CRF (Ours) & \ding{55} & $\mathcal{I}+\mathcal{S}+\mathcal{L}$ & DINO & 77.8 & -- & 53.7 \\

\rowcolor{gray!15}
DS-CRF (Ours, w/ TTA) & \ding{55} & $\mathcal{I}+\mathcal{S}+\mathcal{L}$ & DINO & \textbf{81.1} & \textbf{81.0} & \textbf{56.5} \\

\bottomrule
\end{tabular}
\end{table*}

\begin{table}[t]
\centering
\caption{Comparison with SAM-based WSSS methods. We report mIoU (\%) on the PASCAL VOC 2012 and MS COCO 2014 datasets.}
\label{tab:sam_wsss_comparison}
\small
\begin{tabular}{l c c}
\toprule
Method & VOC \textit{val} & COCO \textit{val} \\
\midrule
\multicolumn{3}{c}{SEPL$_{\text{NeurIPS'23 (Workshop)}}$ \cite{chen_sam_enhances}-enhanced} \\
EPS$_{\text{CVPR'21}}$  \cite{lee_eps} & 72.1 & 41.6 \\
SIPE$_{\text{CVPR'22}}$ \cite{chen_sipe} & 69.7 & 45.2 \\
L2G$_{\text{CVPR'22}}$  \cite{jiang_l2g} & 72.4 & 46.4 \\
CLIMS$_{\text{CVPR'22}}$ \cite{xie_clims} & 71.1 & -- \\
CLIP-ES$_{\text{CVPR'23}}$ \cite{lin_clip_es} & 73.1 & 47.9 \\
\midrule
S2C$_{\text{CVPR'24}}$ \cite{kweon_s2c} & 78.2 & 49.8 \\
Chen and Sun$_{\text{ACM Comput. Surv.'25}}$ \cite{chen_sam_wsss} & 74.0 & 54.6 \\
Sun \textit{et al.}$_{\text{arXiv'23}}$ \cite{sun_alternative} & 77.2 & 55.6 \\
\midrule
\rowcolor{gray!15}
DS-CRF (Ours, w/ TTA) & \textbf{81.1} & \textbf{56.5} \\
\bottomrule
\end{tabular}
\end{table}

\subsection{Datasets and Evaluation Metric}
\label{sec:datasets_and_eval_metric}
We evaluate our framework on the PASCAL VOC 2012 \cite{everingham_pascal} and MS COCO 2014 \cite{lin_ms_coco} datasets. PASCAL VOC 2012 contains 20 foreground classes and one background class, with 1,464 training, 1,449 validation, and 1,456 test images in the original split. 
Following common WSSS practice \cite{kweon_s2c, lin_clip_es}, we instead use the augmented training set of $10,582$ images. MS COCO 2014 includes 80 object categories plus background and is divided into approximately 80k training images and 40k validation images. We report performance using the standard mean Intersection over Union (mIoU) metric.

\subsection{Implementation Details}
\label{sec:implementation_details}
We use DINOv3 (ViT-L/16) \cite{simeoni_dinov3} as our backbone, and generate CAMs with its pre-trained \texttt{dino.txt} head. For SAM, we opt for the lighter ViT-B model for efficiency.

Training is performed using stochastic gradient descent with a fixed learning rate of $1\mathrm{e}{-3}$. The batch size is set to $16$ for PASCAL VOC and $32$ for MS COCO. We train for $10$ epochs on VOC and $4$ epochs on COCO, selecting the checkpoint with the best validation performance. Following \cite{simeoni_dinov3}, we apply image rescaling and horizontal flipping during inference (known as Test-Time Augmentations, or TTA). However, instead of using multiple scales, we find that a single scale at $4\times$ the training resolution works best.

Within our framework, we set the softmax temperature to $\tau = 0.05$ and dilation to $d = 5$. Details on the class-specific weights $\lambda^c$ are provided in the supplementary material. Importantly, tuning all class-specific weights incurs minimal additional cost compared to tuning a single hyperparameter.

Since optimizing our full objective from random weight initializations can lead to degenerate solutions, we first train our model using only the zero-avoiding KL divergence loss $KL(y_i \Vert \sigma_i)$ with soft \texttt{dino.txt} pseudo-labels as targets. This warm-up stage teaches the model to reproduce the soft pseudo-label distributions, which gives a good starting point before we switch to the CRF loss.

\subsection{Comparison to State-of-the-Arts}
\label{sec:comparison_to_sota}
As shown in \cref{tab:sota}, our approach outperforms prior single-stage and multi-stage WSSS methods. Notably, it is the only method in the comparison that does not rely on DenseCRF (dCRF) \cite{krahenbuhl_densecrf}. It therefore avoids dCRF's low-level, colour-based regularization. It also sidesteps the drawbacks of using dCRF to post-process CAMs or final segmentations, as discussed earlier. We further compare against recent SAM-based methods in \cref{tab:sam_wsss_comparison}. Here, DS-CRF again achieves the best performance, showing that our CRF loss is an effective and principled way to use SAM. 

To control for the backbone, we also train the same DINOv3 segmentation network using cross-entropy on pseudo-labels from relevant SAM baselines. Details are provided in the supplementary material.

\subsection{Ablation Studies}

\begin{figure}[t]
\centering
  \captionof{table}{Ablation study of cross-entropy variants and min-max scaling. We report mIoU (\%) on the PASCAL VOC val set.}
  \label{tab:unary_ablation}
  \small
  \begin{tabular}{lccc}
      \toprule
      & \multicolumn{2}{c}{CE} & CCE \\
      \cmidrule(lr){2-3} \cmidrule(lr){4-4}
      & Hard & Soft & Soft \\
      \midrule
      w/o min-max & 40.2 & 43.3 & 74.6 \\
      w/ min-max  & -- & 53.5 & 77.8 \\
      \bottomrule
  \end{tabular}

\vspace{\floatsep}

   \includegraphics[width=0.8\linewidth, trim=15.3bp 22.5bp 15.3bp 15.3bp, clip]{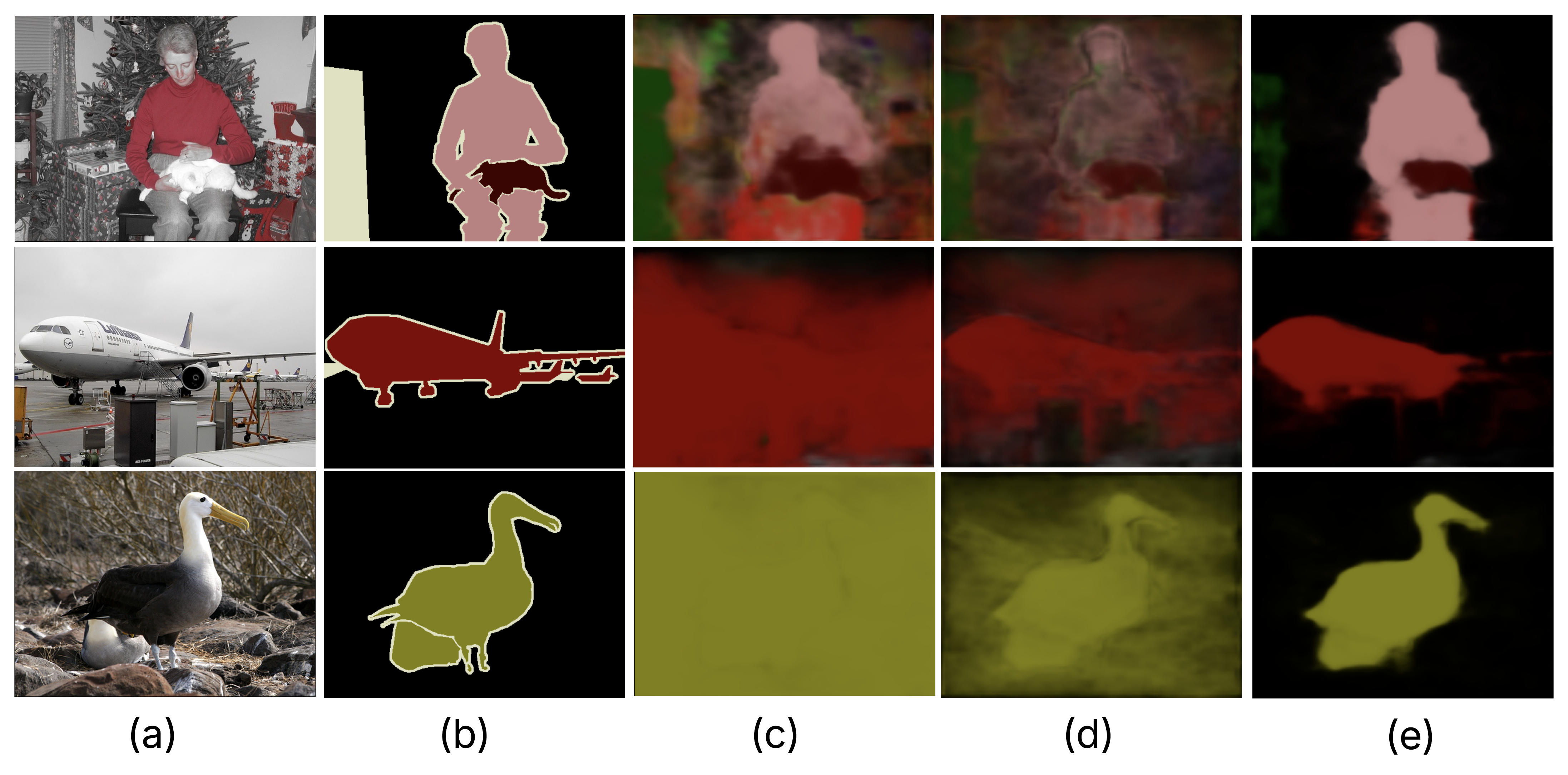}
   \captionof{figure}{Network predictions using different cross-entropy variants. (a) Image. (b) Ground truth. (c) Standard cross-entropy w/ hard pseudo-labels. (d) Standard cross-entropy w/ soft pseudo-labels after min-max scaling. (e) Collision cross-entropy w/ soft pseudo-labels after min-max scaling.}
   \label{fig:unary_ablation}
\end{figure}
\paragraph{Cross-Entropy and Min-Max Scaling.}
We ablate standard versus collision cross-entropy and min-max scaling in \cref{tab:unary_ablation}, keeping pairwise regularization hyperparameters fixed. When supervised by \texttt{dino.txt} softmax outputs (without min-max scaling), collision cross-entropy with soft pseudo-labels markedly outperforms standard cross-entropy with both hard (argmaxed) pseudo-labels and soft pseudo-labels. Applying min-max scaling to soft pseudo-labels adds another $\approx 3\%$ boost to the results. We visualize model predictions using different cross-entropy variants in \cref{fig:unary_ablation}. 
The differences are most pronounced for images dominated by a single large object. In such cases, standard cross-entropy with hard pseudo-labels produces confident but over-expanded predictions, while standard cross-entropy with soft pseudo-labels yields more regularized but uncertain ones.
In contrast, collision cross-entropy achieves a balance, producing predictions that are both regularized and confident.

\begin{figure}[t]
\centering
\captionof{table}{Ablation study of pairwise affinity designs. We report mIoU (\%) on the PASCAL VOC val set.}
\label{tab:pairwise_ablation}
\small
\setlength{\tabcolsep}{1.9pt}
\begin{tabular}{@{}cccccc@{}}
\toprule
Gaussian Kernel & Superpixel & SAM & SAM$^{d=3}$ & SAM$^{d=5}$ & SAM$^{d=7}$ \\
\midrule
50.8 & 51.6 & 50.6 & 74.9 & 77.8 & 75.8 \\
\bottomrule
\end{tabular}

\vspace{\floatsep}

   \includegraphics[width=0.9\linewidth, trim=15.3bp 23.4bp 13.95bp 15.3bp, clip]{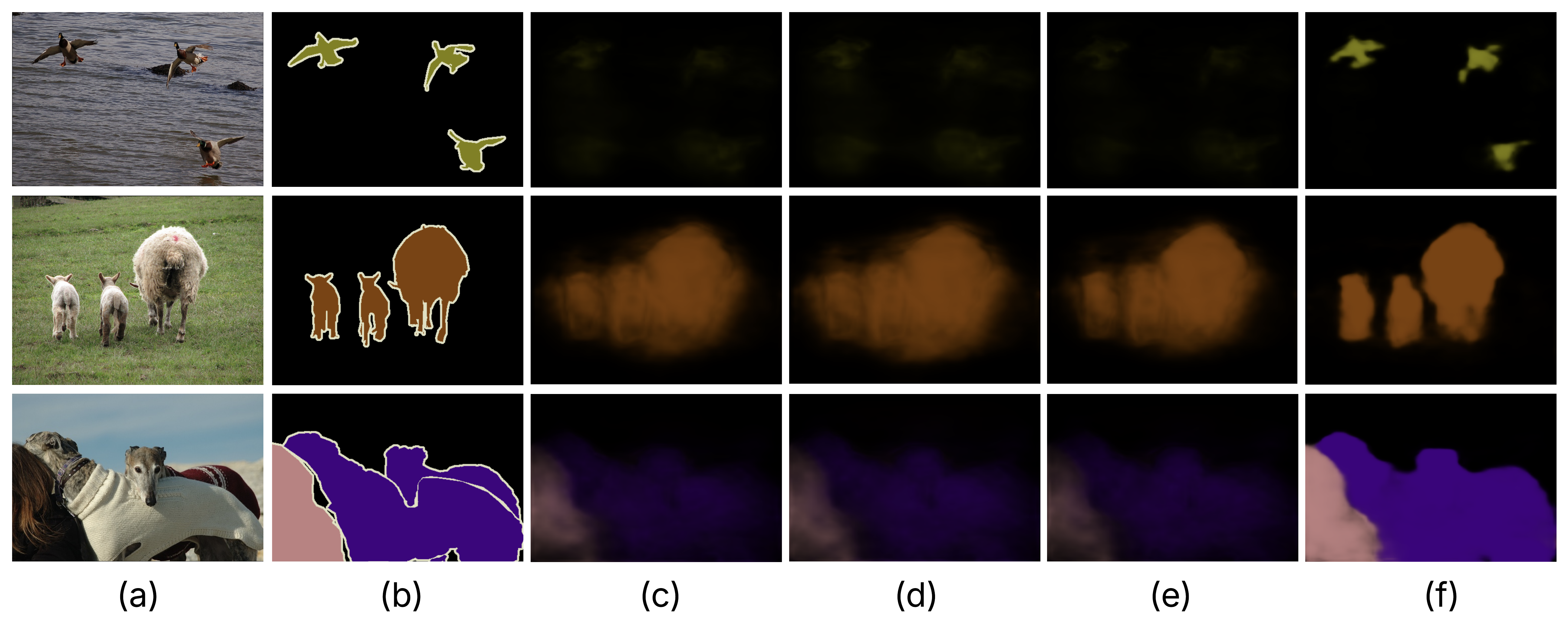}
   \captionof{figure}{Network predictions using different pairwise affinity designs. (a) Image. (b) Ground truth. (c) Gaussian kernel. (d) Superpixel. (e) SAM. (f) SAM with dilation $d=5$.}
   \label{fig:pairwise_ablation}
\end{figure}
\paragraph{Pairwise Affinity.} 
In \cref{tab:pairwise_ablation}, we ablate different pairwise affinity designs in our regularization term while keeping the class-specific weights fixed. With the naive implementation, low-level affinities (e.g. Gaussian kernel on colour differences \eqref{eq:pairwise_gaussian} and SLIC \cite{achanta_slic} superpixel boundaries) perform on par with high-level SAM boundaries. Dilating the SAM boundaries, however, substantially improves performance, peaking at $77.8\%$ mIoU with $d=5$. Low-level affinities do not see the same benefit, since our dilation scheme does not apply to Gaussian kernels and dilating SLIC boundaries reaches only $64.7\%$ mIoU at $d=5$ (not shown in the table). This highlights the value of high-level semantic boundaries in our framework. We also visualize model predictions when trained on these affinities in \cref{fig:pairwise_ablation}. Qualitatively, dilating SAM boundaries is the only design that produces confident predictions with clearly defined boundaries.

\section{Conclusion}

In this paper, we proposed a CRF loss for supervising WSSS training. We realized this loss in DS-CRF, which builds its unary loss term from \texttt{dino.txt} CAMs and its pairwise loss term from SAM boundaries. By keeping these two forms of supervision as separate terms in the loss, our approach departs from prior methods that fuse CAMs and boundaries into a single hard target, which can propagate CAM errors in both size and confidence. Our method also exploits high-level semantic boundaries, rather than the low-level colour cues that DenseCRF relies on. We additionally build our framework from two key insights. First, collision cross-entropy outperforms standard cross-entropy because it lets pseudo-label uncertainty determine the strength of supervision, while still encouraging confident predictions. Second, dilating pairwise affinities is crucial for addressing the partial voluming effect at object boundaries. DS-CRF achieves state-of-the-art performance among single-stage methods, including a new best of 56.5\% mIoU on MS COCO.




{
    \small
    \bibliographystyle{ieeenat_fullname}
    \bibliography{main}
}


\end{document}